\pdfoutput=1
\documentclass[11pt]{article}

\usepackage{acl}

\usepackage{times}
\usepackage{latexsym}
\usepackage{algorithm}
\usepackage{algorithmic}
\usepackage{multirow}
\usepackage{multicol}
\usepackage{makecell}
\usepackage{booktabs}
\usepackage{float}
\usepackage{afterpage}
\usepackage{subcaption}
\usepackage{xcolor}
\usepackage{graphicx}
\usepackage{arydshln}

\usepackage[T1]{fontenc}
\usepackage[utf8]{inputenc}

\usepackage{microtype}

\usepackage{inconsolata}

\title{Metacognitive Prompting Improves Understanding \\ in Large Language Models}

\author{Yuqing Wang \\
  Stanford University \\
  \texttt{ywang216@stanford.edu} \\\And
  Yun Zhao \\
  Meta Platforms, Inc. \\
  \texttt{yunzhao20@meta.com} \\}

\begin{document}
\maketitle
\begin{abstract}
In Large Language Models (LLMs), there have been consistent advancements in task-specific performance, largely influenced by effective prompt design. Recent advancements in prompting have enhanced reasoning in logic-intensive tasks for LLMs, yet the nuanced understanding abilities of these models, crucial for processing and interpreting complex information, remain underexplored. In this study, we introduce \textit{Metacognitive Prompting} (MP), a strategy inspired by human introspective reasoning processes. Using MP, LLMs undergo a systematic series of structured, self-aware evaluations, drawing on both their vast inherent knowledge and new insights. We conduct extensive experiments on four prevalent LLMs: Llama2, PaLM2, GPT-3.5, and GPT-4, across ten natural language understanding (NLU) datasets from GLUE, SuperGLUE, BLUE, and LexGLUE benchmarks. Additionally, we compare our method with chain-of-thought prompting and its advanced versions. The results show that GPT-4 consistently excels across all tasks, while other models have shown significant progress in some tasks when used in conjunction with MP. Furthermore, MP consistently outperforms existing prompting methods in both general and domain-specific NLU tasks. This study underscores the potential to amplify the understanding abilities of LLMs and highlights the benefits of mirroring human introspective reasoning in NLU tasks. Our
data and code are available at \url{https://github.com/EternityYW/Metacognitive-Prompting}.
\end{abstract}

\section{Introduction}
Large Language Models (LLMs) have made significant advancements in natural language processing (NLP) in recent years~\citep{min2021recent, zhao2023survey, wang2023large}. However, as these models progress, simply increasing their scale does not necessarily enhance their understanding and reasoning capabilities~\citep{rae2021scaling}. Delving into the intricacies of prompt design has emerged as a promising approach; it not only rivals the benefits of extensive fine-tuning but also offers clear advantages in sample efficiency~\citep{liu2023pre, kojima2022large}.

Many research efforts have extensively explored prompt design, particularly emphasizing the use of Chain-of-Thought (CoT)~\citep{wei2022chain} approaches to advance intermediate reasoning steps. This led to variants such as Least-to-Most~\citep{zhou2022least}, Self-consistency~\citep{wang2022self}, and  Tree-of-Thoughts (ToT)~\citep{yao2023tree} techniques. These strategies are effective in designated contexts where the main objective centers around enhancing explicit reasoning capacities in areas like arithmetic, commonsense, and symbolic reasoning, guiding LLMs through a logical progression of thought. However, their effectiveness in deepening understanding is limited, as reasoning involves methodically connecting concepts, whereas understanding requires grasping underlying semantics and broader contextual meanings.

\begin{figure}[htbp]
\centering
\includegraphics[width=0.49\textwidth]{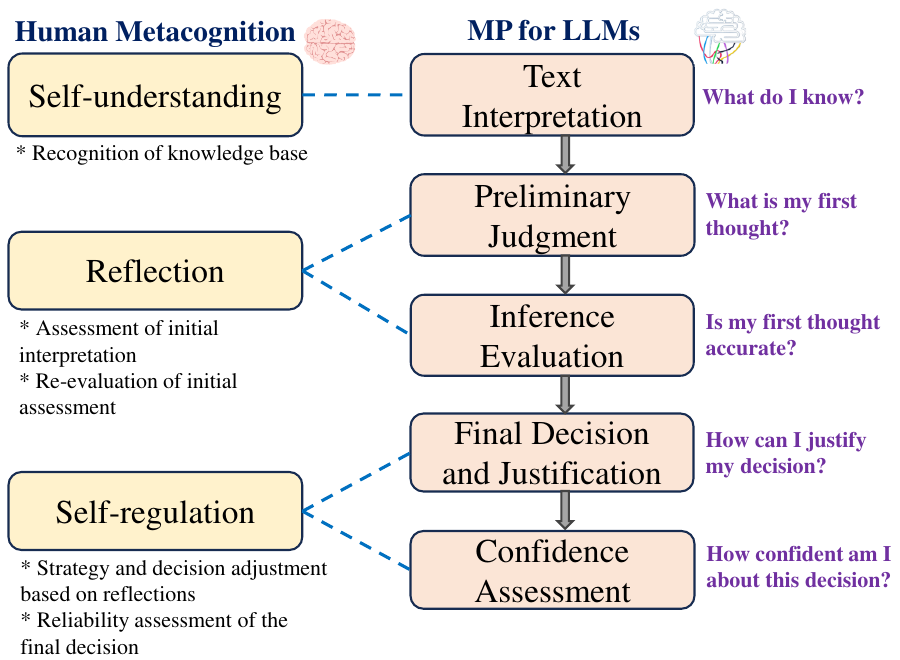}  
\caption{Alignment between human metacognitive processes and the stages of MP in LLMs.}
\label{fig: human_LLM_metacognition}
\end{figure}

To bridge the gap in enhancing LLMs' understanding abilities, crucial for solving complex tasks, we propose Metacognitive Prompting (MP). This method is informed by the concept of metacognition, often defined as ‘thinking about thinking'. Derived from cognitive psychology, metacognition relates to an individual's awareness and self-reflection on their cognitive processes. Our approach integrates key aspects of human metacognitive processes into LLMs. Figure~\ref{fig: human_LLM_metacognition} shows the parallels between human metacognitive stages and the operational steps of our method in LLMs. Rather than concentrating solely on the mechanics of “how” a response is produced, this method delves deeper into the rationale or “why” behind it. The method proceeds as follows: 1) the LLM interprets the provided text, a phase reminiscent of human comprehension; 2) the model then forms an initial judgment, mirroring the stage in which humans generate judgments based on information; 3) the LLM subjects its preliminary inference to critical evaluation, a step aligned with the self-reflection that humans engage in during cognitive processes; 4) after this introspective assessment, the model finalizes its decision and elucidates its reasoning, similar to human decision-making and rationalization; 5) finally, the LLM gauges its confidence in the outcomes, reflecting how humans evaluate the credibility of their judgments and explanations. This paradigm elevates the model's function beyond simple systematic reasoning, compelling it to participate in introspective evaluations that determine the depth and relevance of its responses. 

We conducted experiments on ten NLU datasets from GLUE~\citep{wang2019glue}, SuperGLUE~\citep{wang2019superglue}, BLUE~\citep{peng2019transfer}, and LexGLUE~\citep{chalkidis2022lexglue} benchmarks using several leading LLMs, including Llama2~\citep{touvron2023llama}, PaLM2~\citep{anil2023palm}, GPT-3.5, and GPT-4~\citep{openai2023gpt4}. Our empirical evaluations underscore the superiority of MP over existing prompting strategies, including CoT and its variants. This work emphasizes the importance of incorporating human-inspired introspective reasoning into LLMs, shedding light on an approach that deepens their understanding abilities.

In summary, our contributions are threefold:
\begin{enumerate}
\item[(1)] We introduce \textit{metacognitive prompting}, a novel prompting strategy for LLMs, inspired by human introspective reasoning. This approach formalizes the self-aware evaluation process within LLMs, highlighting the shift from mere task execution to more profound comprehension.
\item[(2)] Our comprehensive experiments on ten NLU datasets reveal that MP outperforms CoT and its variants in both zero-shot and few-shot learning settings. This underscores MP's effectiveness in enhancing the understanding abilities of LLMs.
\item[(3)] Through manual error and confidence analysis, we highlight specific understanding challenges in LLMs. We also illustrate future directions for incorporating human-inspired introspection into LLM comprehension, thereby contributing to enhanced model reliability.
\end{enumerate}

\section{Related Work}
Our proposal for metacognitive prompting is informed by several foundational trajectories: the evolving paradigms of prompting within LLMs, advancements in NLU in the broader NLP domain, and the intricate interplay between cognitive processes and NLU dynamics.
\subsection{Prompting Techniques in LLMs}
Prompts are crucial for harnessing the vast capabilities of LLMs, guiding them to generate accurate outputs or perform specific tasks. Current research primarily focuses on enhancing the reasoning abilities of LLMs. Representative approaches include CoT~\citep{wei2022chain} and its variants like self-consistency~\citep{wang2022self}, Least-to-Most~\citep{zhou2022least}, ToT~\citep{yao2023tree}, and Plan-and-Solve prompting~\citep{wang2023plan}. Additional methods are detailed in~\citep{qiao2022reasoning}. However, there still exists a significant gap in developing effective prompts to enhance NLU within LLMs. Inspired by human cognitive processes, we introduce MP, an approach that not only aims to bridge the understanding gap but also enhances deeper comprehension and reliability in model outputs.

\begin{figure*}[t]
\centering
\includegraphics[width=\textwidth]{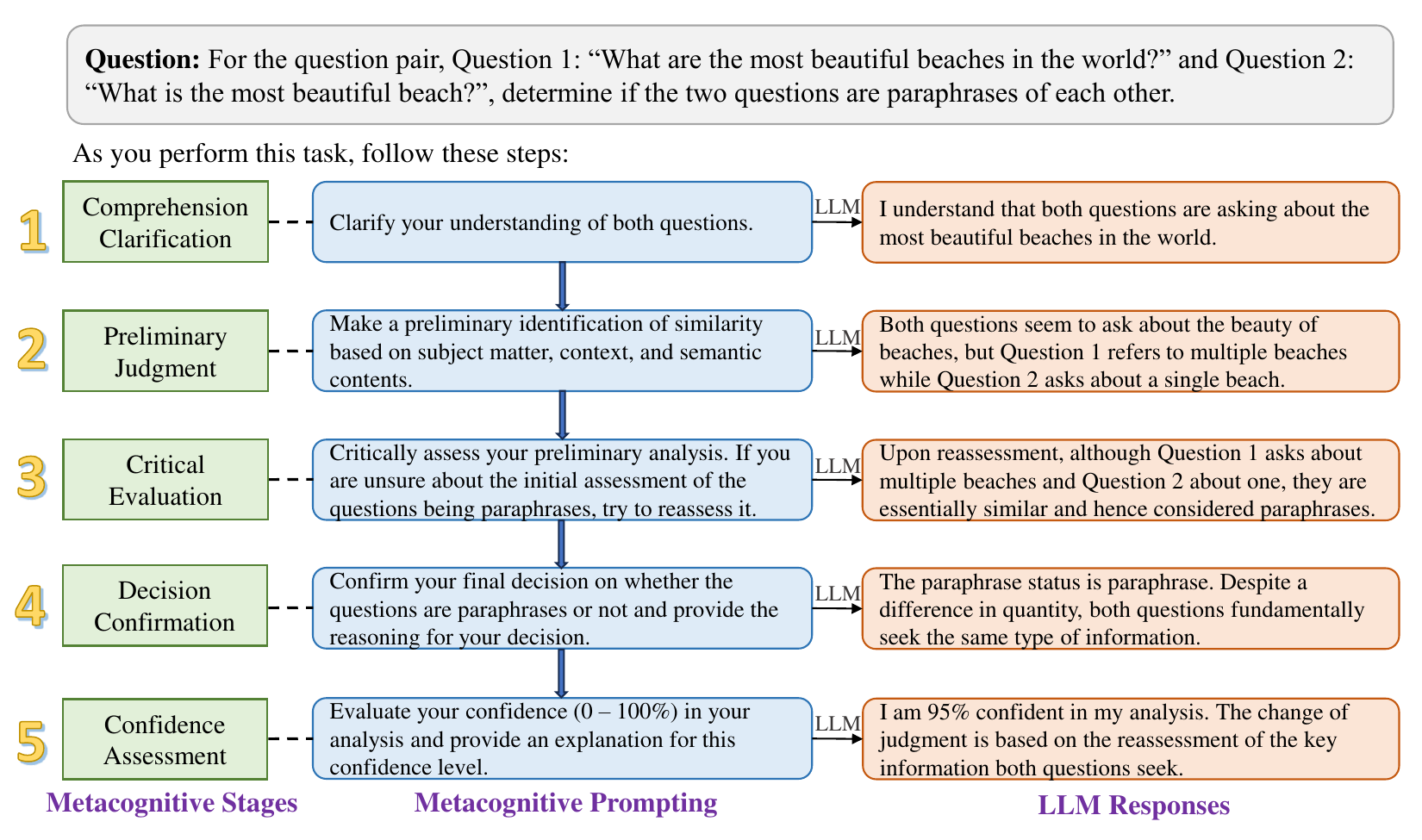}  
\caption{Our proposed method, metacognitive prompting, emulates critical steps of human metacognition, consisting of five stages: 1) understanding the input text, 2) making a preliminary judgment, 3) critically evaluating this preliminary analysis, 4) reaching a final decision accompanied by an explanation of the reasoning, and 5) evaluating the confidence level in the entire process. By reflecting on human self-assessment, these stages guide the LLM, aiding in more accurate text interpretation and facilitating better judgment formation. The diagram features three columns, from left to right, representing the high-level metacognitive stages, specific metacognitive prompts fed into the LLM, and the LLM's corresponding outputs. Prompts in the middle column are collectively fed into the LLM as a single input during the experiments. The figure illustrates a sample question chosen from the Quora Question Pair (QQP) dataset in the GLUE benchmark.}
\label{fig: general_metacognitive_prompting}
\end{figure*}

\subsection{Natural Language Understanding in NLP}
NLU is a fundamental aspect of NLP, emphasizing a model's capacity to grasp the semantics and nuances of human language. Its applications span diverse domains such as question answering (QA)~\citep{namazifar2021language}, text classification~\citep{wang2022integrating, wang2023prominet}, and natural language inference (NLI)~\citep{nie2020can}, as well as commercial tools like chatbots~\citep{ait2020kbot}, voice assistants~\citep{bellegarda2013spoken}, and machine translation. While LLMs have gained remarkable attention recently, with increased efforts dedicated to expanding NLU boundaries, the primary research emphasis has been on their reasoning abilities~\citep{huang2022towards}, ethical use~\citep{weidinger2021ethical, zhuo2023exploring}, and broad applications~\citep{zhao2021empirical, surameery2023use, wang2023empirical}. However, the inherent NLU competencies of LLMs have remained relatively inadequately explored. To address this gap, our study delves into the understanding abilities of various LLMs, employing effective prompting techniques.

\subsection{Cognitive Processes in NLU}
The interplay between cognitive processes and NLU has always been a central consideration in computational linguistics~\citep{perinan2007cognitive, hausser2001foundations}. Cognitive processes, which encompass areas like attention, memory, reasoning, and problem-solving, govern how humans understand, produce, and engage with language in diverse scenarios. These processes heavily influence our linguistic abilities~\citep{allen1995natural, cambria2014jumping}. In the domain of NLU, incorporating cognitive insights may offer improvements in model comprehension. Recognizing this intrinsic connection, our work is inspired to employ a metacognition-based prompting technique, a method rooted in higher-order cognition that reflects on thinking and decision-making, to bolster the understanding capabilities of LLMs, thereby harmonizing traditional modeling techniques with cognitive nuances.

\section{Metacognitive Prompting}

In the complex terrain of human cognition, our ability to introspect and regulate our thinking processes stands as a keystone for intricate problem-solving and decision-making. This high-level cognition underlies our proficiency in breaking down abstract concepts, critically evaluating scenarios, and fine-tuning our reasoning. The primary aim of this work is to equip LLMs with a process that simulates the self-reflective cognitive process. In doing so, we aim to improve LLMs' capabilities in interpreting and responding to NLU tasks.

We propose MP, which instills critical elements of human metacognition into LLMs. This approach involves five distinct stages: 1) the LLM begins by deciphering the input text to comprehend its context and meaning, mirroring the initial comprehension stage in human thought; 2) it then forms a preliminary interpretation of the text, a step that reflects judgment formation in humans; 3) subsequently, the LLM critically evaluates this initial judgment for accuracy, akin to the self-scrutiny humans apply during problem-solving; 4) after this evaluation, the LLM finalizes its decision and offers an explanation for its reasoning, aligning with the decision-making and rationalization phase in human cognition; 5) ultimately, the LLM assesses its confidence in the outcome of the entire process, similar to how humans gauge the certainty of their decisions and explanations. Figure~\ref{fig: general_metacognitive_prompting}  provides a schematic representation of our MP. It outlines the five sequential metacognitive stages, the specific prompts directed at the LLM, and corresponding model outputs.

%\begin{table*}[htbp]
%\centering 
 % \caption{Comparison among standard prompting, chain-of-thought prompting, and metacognitive prompting.}
 % \begin{tabular}{ccc}
 %   \toprule
% \textbf{Prompting Strategy} & \textbf{Guideline} & \textbf{Purpose} \\ \midrule 
% Standard & \makecell{Provide a straightforward, \\ succinct prompt tailored \\ to the task at hand.} & \makecell{To elicit an immediate \\ and direct answer from \\ the model.} \\  
% \midrule 
% Chain-of-Thought & \makecell{Develop a prompt that \\ leads the model through \\ progressive stages of reasoning.} & \makecell{To facilitate the model's \\ step-by-step logical \\ engagement with tasks.} \\ 
% \midrule 
% Metacognitive & \makecell{Formulate a prompt that \\ simulate stages of \\ human metacognition.} & \makecell{To reproduce human-like  \\ “thinking about thinking", \\ deepening the model's \\ task comprehension.}\\
%  \bottomrule
%\end{tabular}
%\label{tab: comparison_prompting}
%\end{table*}

\begin{table*}[h]
\centering 
\caption{Overview of NLU datasets belong evaluated. \textit{WSD} stands for word sense disambiguation, \textit{NER} for named entity recognition, \textit{RE} for relation extraction, \textit{MLC} for multi-label classification, and \textit{MCC} for multi-class classification. Acc., $\mu$-F1 and m-F1 represent accuracy, micro-F1 and macro-F1, respectively.}
\resizebox{\linewidth}{!}{%
\begin{tabular}{cccccc}
    \toprule
\textbf{Source Benchmark} & \textbf{Dataset} & \textbf{Task} & \textbf{\# Classes} & \textbf{Metrics} & \textbf{Domain} \\ 
    \midrule
\multirow{2}{*}{\textbf{GLUE}} & QQP & Paraphrase & 2 (paraphrase or not) & acc./F1 & Social QA \\
& QNLI & QA/NLI & 2 (entailment or not) & acc. & Wikipedia \\
\midrule
\multirow{2}{*}{\textbf{SuperGLUE}} & BoolQ & QA & 2 (yes/no) & acc. & Wikipedia, Google queries\\
& WiC & WSD & 2 (True/False) & acc. & WordNet, Wiktionary, etc. \\
\midrule
\multirow{3}{*}{\textbf{BLUE}} & BC5CDR-chem & NER & 3 (BIO tags) & $\mu$-F1 & Biochemistry \\
& DDI & RE & 4 (Advice, Effect, etc.) & m-F1 & Biochemistry \\
& MedNLI & NLI & 3 (ECN relations) & acc. & Clinical practice \\
\midrule
\multirow{3}{*}{\textbf{LexGLUE}} & EUR-LEX & MLC & 100 (EuroVoc concepts) & $\mu$-F1/m-F1 & EU Law \\
& LEDGAR & MCC & 100 (contract provisions) & $\mu$-F1/m-F1 & Contracts \\
& UNFAIR-ToS & MLC & 8 + 1 (unfair terms) & $\mu$-F1/m-F1 & Contracts \\
\bottomrule
\end{tabular}}
\label{tab: dataset}
\end{table*}

%Table~\ref{tab: comparison_prompting} contrasts our MP with prevalent prompting methods such as standard prompting (SP) and CoT prompting, highlighting the differences in guidelines and objectives inherent to each method. While SP focuses on direct, task-specific cues, MP uniformly applies the metacognitive process, adapting to the unique demands of every task. Contrary to the sequential progression characteristic of CoT, MP integrates continuous critical evaluations throughout its stages, enhancing both comprehension and response. For instance, in a sentiment analysis task, SP might simply request, “Classify the sentiment of the statement as positive or negative.” Meanwhile, CoT guides the model through a step-by-step process, asking, “Identify key emotional words in the statement. Based on these words, would you classify its overall sentiment as positive or negative?" On the other hand, MP pushes the model for deeper introspection, suggesting, “Understand the statement and make a preliminary sentiment identification. If you are uncertain, reassess. Confirm your final decision, providing reasoning. Then, evaluate and justify your confidence (0 - 100\%) in this analysis."

In essence, MP introduces a structured approach that enables LLMs to process tasks, enhancing their contextual awareness and introspection in responses. By systematically guiding models through stages that emulate human cognitive processes, this method offers a fresh perspective on addressing complex natural language tasks. It reshapes our perception and utilization of LLMs' capabilities, ushering in a paradigm where models not only grasp the intricacies of given tasks but also critically evaluate and adjust their responses. This approach establishes a foundation for more effective and reliable interactions between users and LLMs, particularly benefiting those with limited LLM expertise, as it simplifies complex linguistic and cognitive processes into more manageable forms. Sample MP templates and exemplars are shown in Appendix~\ref{sample_prompts}.

\section{Experiments}
We conduct experiments on ten diverse NLU datasets selected from GLUE~\citep{wang2019glue}, SuperGLUE~\citep{wang2019superglue}, BLUE~\citep{peng2019transfer}, and LexGLUE~\citep{chalkidis2022lexglue} benchmarks. We evaluate the impact of MP in comparison with CoT and its variants, across four leading LLMs. We report the best result after multiple experimental iterations. 

\subsection{Datasets}
For our experiments, we use a broad set of datasets from the GLUE, SuperGLUE, BLUE, and LexGLUE benchmarks, encompassing both general NLU and domain-specific datasets in biomedicine and law. In general NLU, our selections include question paraphrase (QQP~\citep{shankar2017first}), question-answer entailment (QNLI~\citep{rajpurkar2016squad}), QA (BoolQ~\citep{clark2019boolq}), and word sense disambiguation (WiC~\citep{pilehvar2019wic}). For biomedical NLU, we select named entity recognition (BC5CDR-chem~\citep{li2016biocreative}), relation extraction (DDI~\citep{segura2013semeval}), and NLI (MedNLI~\citep{romanov2018lessons}). For legal NLU, we opt for multi-label text classification (EUR-LEX~\citep{chalkidis2021multieurlex}, UNFAIR-ToS~\citep{lippi2019claudette}) and multi-class text classification (LEDGAR~\citep{tuggener2020ledgar}). These datasets pose diverse challenges to the understanding abilities of LLMs. Given the constraints of API costs, we randomly select 600 examples from the validation set of each dataset. Table~\ref{tab: dataset} provides an overview of the tasks and datasets.

\subsection{Prompts} Our proposed MP is adaptable to both zero-shot and 5-shot settings. 
For each setting, we consider the following prompting baselines: (1) Zero-shot CoT~\citep{kojima2022large}, which adds “\textit{Let’s think step by step}” to a basic query, and Plan-and-Solve (PS) prompting~\citep{wang2023plan}, which appends “\textit{Let’s first understand the problem and devise a plan to solve the problem. Then, let’s carry out the plan and solve the problem step by step}” to the end of a question, are included as zero-shot baselines. (2) Manual-CoT~\citep{wei2022chain} and self-consistency with CoT (CoT-SC)~\citep{wang2022self}, the latter of which takes majority vote from 10 CoT samples, are considered as few-shot baselines. Exemplars for each dataset are hand-crafted.

\subsection{Large Language Models} In our evaluation, we consider four popular LLMs: the open-source model Llama-2-13b-chat~\citep{touvron2023llama} and the closed-source models PaLM-bison-chat~\citep{anil2023palm}, GPT-3.5-turbo, and GPT-4~\citep{openai2023gpt4}. Each model is employed using its corresponding API key. For all methods, we apply greedy decoding (i.e., temperature = 0) for response generation, except when applying CoT-SC (temperature = 0.7). Furthermore, we utilize zero-shot and 5-shot settings for each model, with exemplars for the 5-shot setting randomly selected from the training set. Each dataset has its unique set of exemplars, and the answers for these exemplars are obtained through human annotation.

\begin{table*}[h]
\centering 
\caption{Performance comparison of four LLMs across ten NLU datasets. The best results for the 5-shot setting (5S) are boldfaced, and for the zero-shot setting (0S), underlined. M-CoT and M-MP indicate manually-designed demonstrations in the 5-shot setting. GPT-4 consistently outperforms other models across all NLU datasets. MP notably surpasses other prompting baselines in the majority of tasks.}
\resizebox{\linewidth}{!}{%
\renewcommand{\arraystretch}{1.15}
\begin{tabular}{lcccccccccc}
 \toprule
  \multirow{3}{*}{\textbf{Method}} & & & & & &\textbf{Dataset} \\ \cmidrule(lr){2-11} 
    &  QQP & QNLI & BoolQ & WiC & BC5CDR-chem & DDI & MedNLI & EUR-LEX & LEDGAR & UNFAIR-ToS\\ 
    &  \textit{acc./F1} & \textit{acc.} & \textit{acc.} & \textit{acc.} & \textit{$\mu$-F1} & \textit{m-F1} & \textit{acc.} & \textit{$\mu$-F1/m-F1} & \textit{$\mu$-F1/m-F1} & \textit{$\mu$-F1/m-F1} \\ \midrule
     Llama2 (0S, CoT) & 84.5/79.5 & 89.5 & 81.9 & 75.2 & 94.2 & 70.5 & 58.3 & 25.6/14.5 & 60.8/47.6 & 43.9/26.7\\
     Llama2 (0S, PS) & 85.6/80.8 & 89.9 & 83.1 & 76.0 & 95.6 & 72.0 & 59.1 & 27.8/16.9 & 61.4/48.1 & 46.1/28.4\\
     Llama2 (0S, MP) & 86.9/82.1 & 90.4 & 86.3 & 78.8 & 96.0 & 74.3 & 62.8 & 32.5/21.4 & 63.8/50.5 & 50.2/31.6\\
     \hdashline
     PaLM2 (0S, CoT) & 85.4/80.6 & 89.9 & 88.1 & 76.4 & 94.5 & 70.9 & 61.1 & 24.8/13.1 & 63.9/49.1 & 46.2/29.1\\
     PaLM2 (0S, PS) & 85.2/80.3 & 89.5 & 89.5 & 77.1 & 94.9 & 72.8 & 60.9 & 26.1/14.8 & 65.0/52.7 & 47.4/30.8\\
     PaLM2 (0S, MP) & 86.2/81.9 & 90.8 & 90.5 & 78.8 & 96.2 & 74.0 & 63.3 & 29.3/16.5 & 67.6/54.8 & 52.5/33.7\\ 
     \hdashline
     GPT-3.5 (0S, CoT) & 84.9/79.9 & 90.3 & 84.8 & 76.9 & 93.9 & 63.9 & 70.6 & 31.9/20.7 & 68.1/57.6 & 50.4/33.2\\
     GPT-3.5 (0S, PS) & 84.7/80.6 & 90.8 & 85.0 & 76.6 & 94.2 & 66.1 & 72.3 & 33.6/21.8 & 68.9/58.3 & 52.3/34.8\\
     GPT-3.5 (0S, MP) & 86.1/81.5 & 92.3 & 87.7 & 78.4 & 94.8 & 70.7 & 76.4 & 36.7/23.5 & 70.2/59.8 & 56.7/38.1\\
     \hdashline
     GPT-4 (0S, CoT) & 88.9/84.7 & 95.0 & 90.4 & 82.0 & 97.3 & 72.1 & 78.2 & 37.4/24.8 & 73.6/59.4 & 54.7/38.5\\
     GPT-4 (0S, PS) & 89.4/85.3 & 96.2 & 90.7 & 82.4 & 97.6 & 73.5 & 79.8 & 39.6/27.1 & 75.4/60.7 & 58.3/41.7\\
     GPT-4 (0S, MP) & \underline{89.9/86.2} & \underline{97.1} & \underline{91.4} & \underline{83.6} & \underline{98.5} & \underline{74.7} & \underline{81.1} & \underline{43.8/29.9} & \underline{78.1/62.8} & \underline{64.0/45.3}\\
    \midrule
     Llama2 (5S, M-CoT) & 85.2/80.2 & 90.1 & 82.8 & 76.5 & 94.9 & 73.8 & 61.2 & 23.3/12.7 & 54.7/43.3 & 52.8/35.6\\
     Llama2 (5S, CoT-SC) & 86.1/80.9 & 90.8 & 84.2 & 76.9 & 95.3 & 76.2 & 63.5& 24.6/14.7 & 55.6/44.8 & 55.6/37.9\\
     Llama2 (5S, M-MP) & 88.1/83.2 & 91.6 & 87.4 & 79.5 & 96.6 & 77.3 & 64.7 & 27.8/15.9 & 58.2/46.6 & 59.7/41.2\\
     \hdashline
     PaLM2 (5S, M-CoT) & 85.8/81.3 & 90.9 & 89.2 & 77.7 & 95.1 & 73.1 & 63.3 & 22.8/12.0 & 57.5/45.2 & 57.4/31.9\\
     PaLM2 (5S, CoT-SC) & 86.9/81.7 & 91.7 & 90.9 & 78.2 & 96.4 & 75.4 & 63.8 & 23.9/13.8 & 57.9/45.7 & 60.2/34.6\\
     PaLM2 (5S, M-MP) & 87.9/82.5 & 93.8 & 90.9 & 79.6 & 96.2 & 75.2 & 65.1 & 26.7/15.4 & 59.3/47.3 & 65.4/38.8\\
     \hdashline
     GPT-3.5 (5S, M-CoT) & 85.1/80.2 & 91.2 & 86.7 & 77.4 & 94.7 & 67.8 & 74.3 & 29.3/19.5 & 61.7/50.1 & 62.3/45.1\\
     GPT-3.5 (5S, CoT-SC) & 86.1/81.7 & 91.4 & 88.3 & 78.8 & 95.7 & 70.1 & 76.5 & 30.6/19.8 & 63.0/51.4 & 65.7/47.2\\
     GPT-3.5 (5S, M-MP) & 86.4/81.9 & 93.1 & 89.7 & 79.1 & 96.6 & 71.6 & 78.1 & 32.4/20.7 & 64.9/53.7 & 69.1/50.1\\
     \hdashline
     GPT-4 (5S, M-CoT) & 89.5/85.6 & 95.8 & 90.8 & 82.3 & 97.9 & 74.6 & 80.1 & 35.3/22.6 & 66.4/57.2 & 69.2/50.3\\
    GPT-4 (5S, CoT-SC) & 90.1/86.7 & 96.8 & 91.6 & 83.4 & 98.9 & 76.9 & 80.5 & 37.6/24.4 & 68.2/58.4 & 72.8/54.1\\
     GPT-4 (5S, M-MP) & \textbf{91.3/88.2} & \textbf{98.9} & \textbf{92.0} & \textbf{84.3} & \textbf{99.4} & \textbf{80.8} & \textbf{82.4} & \textbf{40.1/28.8} & \textbf{70.3/59.9} & \textbf{75.6/55.8}\\
   \bottomrule
\end{tabular}}
\label{tab: overall_performance}
\end{table*}

\section{Results}
In our empirical evaluations, we compare performance across all datasets and models, considering the various prompting methods used. We also investigate the efficacy of different prompting strategies, analyze errors associated with MP, and examine the relationship between confidence scores and predictive performance when MP is applied.

\subsection{Overall Performance Comparison}
Table~\ref{tab: overall_performance} presents a comprehensive performance comparison of our method against established zero-shot and few-shot methods on four LLMs across ten varied NLU datasets. Generally, 5-shot learning outperforms zero-shot learning across models, except for EUR-LEX and LEDGAR. The latter's performance dip may be attributable to their high-class counts and the limited example demonstrations, which can skew the models toward a narrow label set. Particularly, zero-shot MP outperforms M-CoT in some instances, suggesting that reduced manual effort can still effectively elicit deep understanding in LLMs, potentially inspiring the development of more efficient prompting methods. Furthermore, GPT-4 stands out, consistently scoring highest on all datasets by a significant margin. For zero-shot prompting, LLMs exhibit notably improved performance with MP, particularly for legal NLU tasks like EUR-LEX. Specifically, MP boosts $\mu$-F1 by $15.0\%$ to $26.9\%$ over CoT and by $9.2\%$ to $16.9\%$ over PS on the EHR-LEX dataset. A similar trend is seen with 5-shot methods; for instance, on the same dataset, M-MP enhances $\mu$-F1 by $10.6\%$ to $19.4\%$ over M-CoT and by $5.9\%$ to $13.0\%$ over CoT-SC. Overall, integrating MP yields substantial benefits for domain-specific NLU datasets in the fields of biomedicine and law across all models. It also provides a moderate yet consistent improvement in general NLU tasks.

\subsection{Prompting Strategy Comparison}
We evaluate the performance of different prompting strategies under zero-shot and 5-shot learning settings across all models and datasets. 

\begin{figure}[ht]
     \centering
     \begin{subfigure}[b]{0.45\textwidth}
         \centering
         \includegraphics[width=\textwidth]{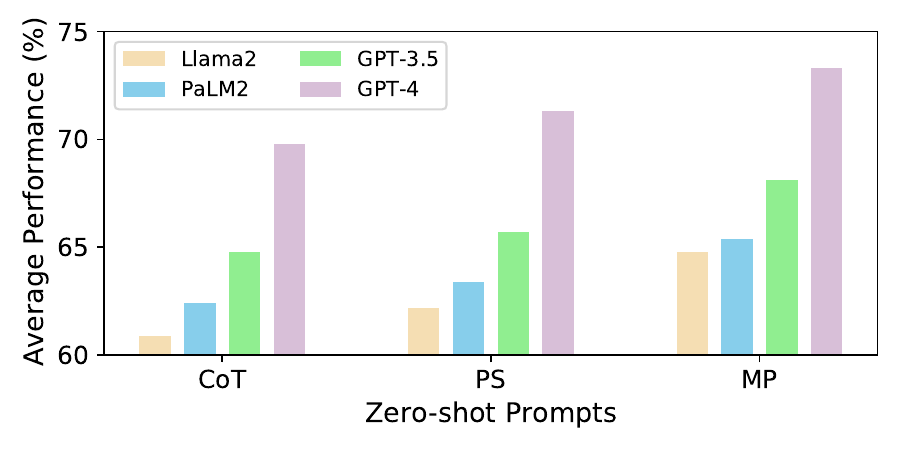}
     \end{subfigure}
     \hfill
     \begin{subfigure}[b]{0.45\textwidth}
         \centering
         \includegraphics[width=\textwidth]{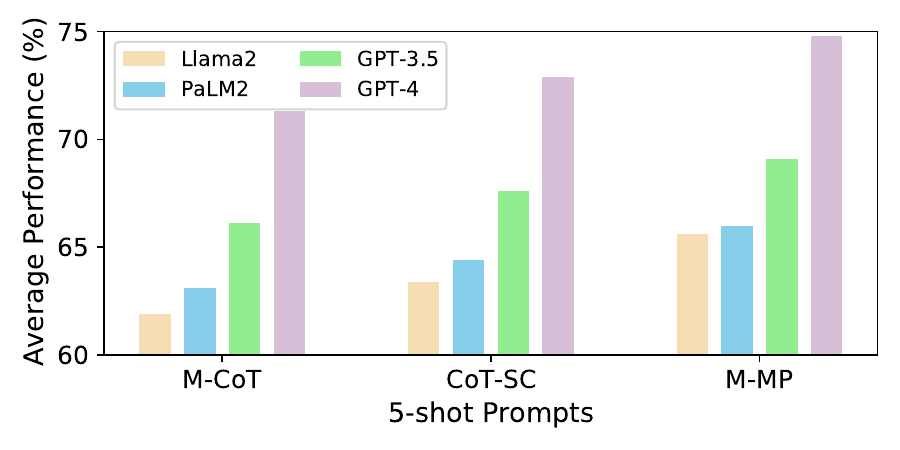}
     \end{subfigure}
        \caption{Comparison of average performance for all prompting methods in both zero-shot and 5-shot learning scenarios across four LLMs. Performance metrics are averaged over all datasets, treating each dataset and metric with equal significance and assuming direct comparability. MP consistently surpasses other methods.}
        \label{fig: prompting_performance_models}
\end{figure}

In the model-level comparison, Figure~\ref{fig: prompting_performance_models} presents an aggregated view of the performance of each prompting method across all datasets for each model (top for zero-shot and bottom for 5-shot), assuming that datasets and evaluation metrics are equally significant and directly comparable. For the zero-shot learning setting, MP emerges as superior, illustrating a relative performance boost ranging from 4.8\% to 6.4\% over CoT and 2.8\% to 4.1\% over PS. Similarly, M-MP shows an average performance improvement from 4.5\% to 6.0\% over M-CoT and 2.2\% to 3.5\% over CoT-SC in the 5-shot learning setting. This enhanced performance can be attributed to the unique introspective strategy of MP, which facilitates a deeper understanding of tasks by prompting the model to critically evaluate, revisit its initial judgments, and refine its responses. When we shift focus to a data-level comparison, considering zero-shot learning results as an example, Table~\ref{tab: prompting_performance_data} provides an average performance over four LLMs for each dataset. The critical reassessment capabilities of MP particularly stand out in datasets like MedNLI, UNFAIR-ToS, and EUR-LEX, leading to marked improvements of 4.3\%, 9.6\%, and 12.4\% over PS (enhanced version of zero-shot CoT), respectively. The consistent outstanding performance of MP underscores its potential in tasks demanding precision, discernment, and a comprehensive semantic grasp. Meanwhile, the self-assessment and iterative refinement embedded in MP give it an advantage in tasks requiring nuanced understanding and contextual depth.

\begin{table}[h]
\centering 
\caption{Comparison of average performance for zero-shot prompting methods across datasets. Performance metrics are averaged over all models. MP consistently achieves superior performance across all NLU tasks.}
\resizebox{\linewidth}{!}{%
\begin{tabular}{lccc}
 \toprule
    \textbf{Dataset} & \textbf{CoT} & \textbf{PS} & \textbf{MP}\\ \midrule 
   QQP (\textit{acc./F1}) & 85.9/81.2 & 86.2/81.7 & \textbf{87.3/82.9}\\
   QNLI (\textit{acc.}) & 91.2 & 91.6 & \textbf{92.6} \\
   BoolQ (\textit{acc.}) & 86.3 & 87.1 & \textbf{89.0} \\
   WiC (\textit{acc.}) & 77.6 & 78.0 & \textbf{79.9}\\
   BC5CDR-chem (\textit{$\mu$-F1}) & 95.0 & 95.6 & \textbf{96.4} \\
   DDI (\textit{m-F1}) & 69.4 & 71.1 & \textbf{73.4}\\
   MedNLI (\textit{acc.}) & 67.1 & 68.0 & \textbf{70.9}\\
   EUR-LEX (\textit{$\mu$-F1/m-F1}) & 29.9/18.3 & 31.8/20.2 & \textbf{35.6/22.8} \\
   LEDGAR (\textit{$\mu$-F1/m-F1}) & 66.6/53.4 & 67.7/54.9 & \textbf{69.9/57.0} \\
   UNFAIR-ToS (\textit{$\mu$-F1/m-F1}) & 48.8/31.9 & 51.0/33.9 & \textbf{55.8/37.2} \\
   \bottomrule
\end{tabular}}
\label{tab: prompting_performance_data}
\end{table}

\begin{figure}[h]
     \centering
     \begin{subfigure}[b]{0.45\textwidth}
         \centering
         \includegraphics[width=\textwidth]{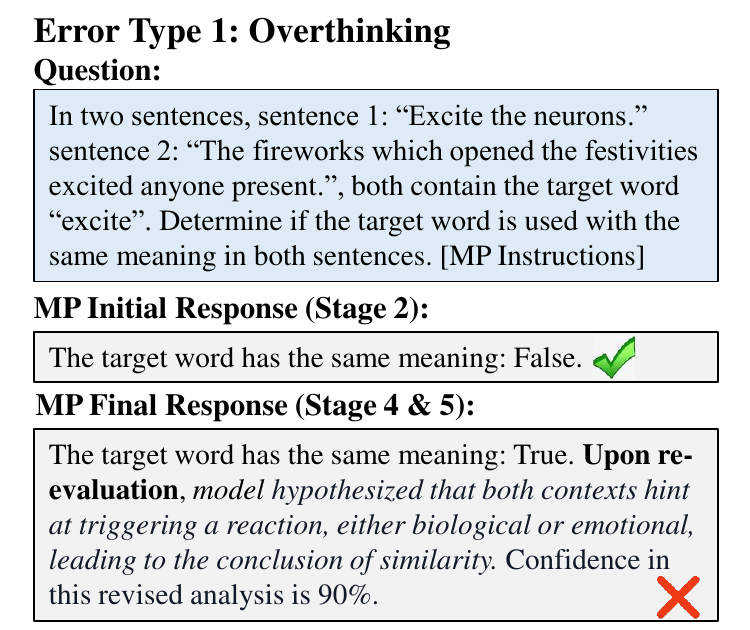}
         \caption{Overthinking error in model response with MP.}
     \end{subfigure}
     \hfill
     \begin{subfigure}[b]{0.45\textwidth}
         \centering
         \includegraphics[width=\textwidth]{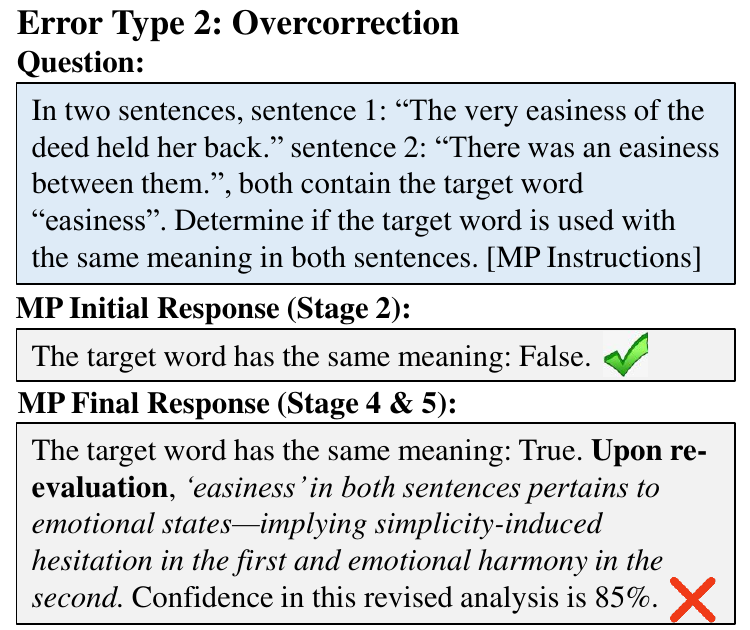}
         \caption{Overcorrection error in model response with MP.}
     \end{subfigure}
        \caption{Two major error types with MP: overthinking (excessive analysis) and overcorrection (excessive adjustment). Example questions are from the WiC dataset.}
        \label{fig: mp_errors}
\end{figure}

\subsection{Error Analysis}
MP has consistently demonstrated proficiency across a range of NLU tasks. However, upon manual inspection of its incorrect predictions, we identify two primary error types across all tasks (10 datasets) specifically associated with MP. First, ‘Overthinking errors’ (68.3\%) are notably evident in straightforward datasets like QQP and BoolQ. In these situations, MP tends to over-complicate the task, diverging from the correct solution. Conversely, ‘Overcorrection errors’ (31.7\%) predominantly appear in tasks demanding nuanced interpretation, such as WiC and DDI. This type of error appears obvious in the critical reassessment stage of MP, which strays excessively from an initially accurate interpretation. Figure~\ref{fig: mp_errors} shows examples of both error types from the WiC dataset. In addition, we observe distinct error patterns in domain-specific tasks. In biomedical NLU tasks (3 datasets), MP predominantly encounters errors including ‘Terminological misalignments’ (48.6\%), where the model inaccurately interprets specialized medical terms, and ‘Clinical inference discrepancies’ (51.4\%), where the depth and interconnections of clinical data are not fully comprehended or are misapplied. In legal NLU tasks (3 datasets), the errors are often characterized as ‘Statutory interpretation errors’ (52.2\%), reflecting challenges in deciphering the complex language and context of legal documents, and ‘Jurisprudential analysis deviations’ (47.8\%), where the model diverges from accepted legal reasoning or misinterprets legal principles and precedents. Numbers in parentheses represent the approximate distributions of major error types within the subgroup. These error types, unique to the specific demands of biomedicine and law, highlight the need for tailored adjustments in MP’s further application to these fields.

\subsection{Confidence Analysis}
Assessing confidence and uncertainty within the MP framework is instrumental in gauging the reliability of predictions, particularly when models articulate their confidence levels. In our analysis, each model operating with MP is evaluated based on its verbalized confidence for every prediction across the datasets. Scores above 75\% are classified as high confidence; any value below this threshold is considered low confidence. To illuminate this correlation, we employ a tailored confusion matrix uniquely adapted for this study. Within this matrix, the standard terminologies of ‘True Positive', ‘False Positive', ‘True Negative', and ‘False Negative' are redefined as follows:

 \noindent \textbf{True Positive (TP):} Represents instances where the model, using MP, expressed high confidence and produced a correct answer. These account for 55.6\%.

\noindent \textbf{False Positives (FP):} Denotes cases where the model exhibited high confidence but gave an incorrect prediction. These amount to 32.5\%.

\noindent \textbf{True Negatives (TN):} Refers to instances where the model signaled low confidence and its response was indeed incorrect. These stand at 6.8\%.

\noindent \textbf{False Negatives (FN):} Highlights cases where the model indicated low confidence but, surprisingly, delivered a correct answer. These tally to 5.1\%.
\begin{figure}[h]
\centering
\includegraphics[width=0.5\textwidth]{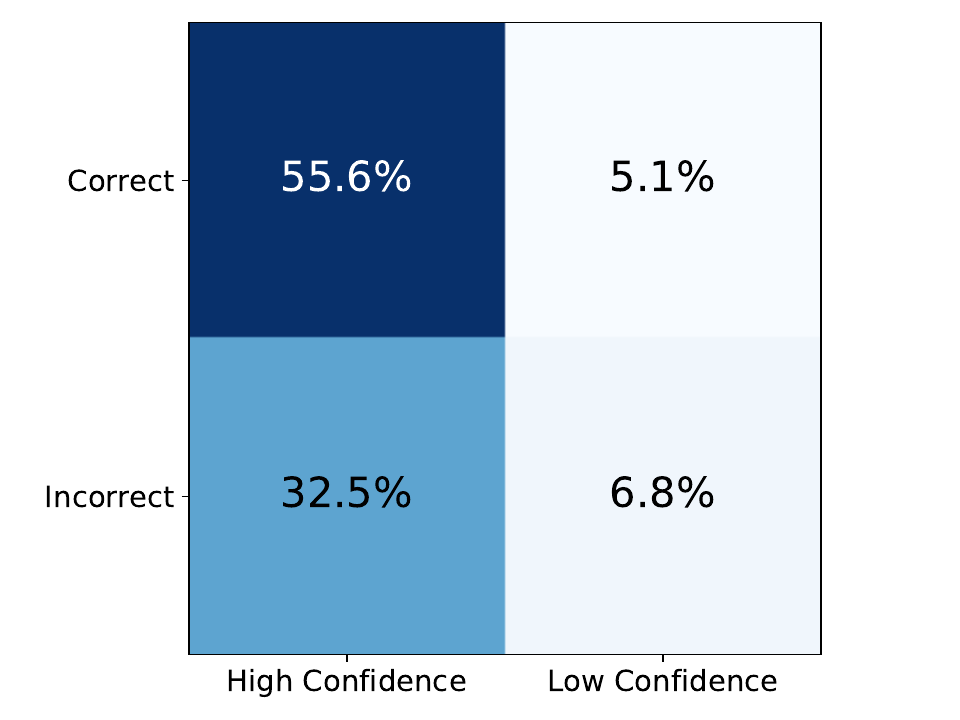}  
\caption{The relationship between correctness and confidence levels under MP, averaged over all datasets and models.}
\label{fig: mp_confidence}
\end{figure}

These metrics are aggregated across all models and datasets and then averaged to provide a holistic overview of the interplay between model confidence using MP and prediction accuracy. As depicted in Figure~\ref{fig: mp_confidence}, MP typically offers an accurate reflection of its own performance, as evidenced by the high TP rate. The relatively low TN rate underscores its reliable self-assessment, suggesting that when MP has low confidence, it is predominantly correct about its inaccuracy. However, the considerable FP rate indicates that, while MP is usually right when confident, it sometimes makes mistakes despite its high confidence. Moreover, the FN rate identifies areas where MP might improve its self-awareness, as there are moments when it might underestimate its accuracy. In summary, the high TP rate and low FN values underscore MP's self-awareness, but the FP and TN values point to potential improvements. Addressing these areas by emphasizing confidence calibration in future iterations of MP could better align its introspective evaluations with its actual performance abilities.

\section{Limitations}
While our proposed MP demonstrates potential by integrating introspective features reminiscent of human cognition into LLMs to enhance their understanding capacities, our study does have its limitations. First, designing the prompts requires manual effort to guide the LLMs through metacognitive processes. Second, we evaluate the effectiveness of MP using a selection of datasets and models, which may limit the broader applicability of our findings. Furthermore, although the verbalized confidence of LLMs offers a window into their perceived certainty levels, it might not serve as the definitive method for comprehensively gauging their true confidence. A hybrid approach, such as combining verbalization with self-consistency checks, could offer a more robust method for confidence calibration. Additionally, our study does not extensively address vital ethical and legal concerns, such as potential biases, privacy implications, and fairness challenges. Future research on MP will address these dimensions to ensure the responsible and holistic application of LLMs in different areas.

\section{Discussion}
In this study, we present MP to infuse introspective features that mirror human cognition into LLMs. The MP process involves five distinct stages: it starts by comprehending the input text, then moves to formulate an initial judgment. Next, it critically reevaluates this initial impression, settles on a decision while explaining its rationale, and finally gauges its confidence in the decisions made. We conduct experiments on a broad range of datasets from several popular NLU benchmarks and evaluate several prominent LLMs with different prompting methods. The results underscore the potential of our method, demonstrating advantages over existing prompting methods. Through our analysis, specific error patterns associated with MP are identified, highlighting nuances in comprehension and judgment stages that warrant further refinement. While MP provides a structured pathway for models to introspect, it follows predefined stages, lacking adaptability based on real-time feedback. The five-stage design of MP, although foundational, suggests room for more intricate frameworks that might emulate human-like cognitive feedback loops more authentically.

Looking forward, several areas warrant further exploration. First, we plan to apply MP more broadly, particularly to detail-oriented areas such as mental health support, as well as to complex reasoning tasks like arithmetic and commonsense reasoning. Refining MP could elicit more detailed introspective responses from LLMs. Moreover, reliance on verbalized confidence can be augmented by integrating other methods for a more comprehensive assessment. Additionally, the broader implications of introducing introspective LLMs, particularly regarding biases and the reliability of outputs, require in-depth examination. In essence, our initial venture with MP lays a solid foundation, but significant opportunities remain to draw closer parallels between introspection in LLMs and natural human introspection, which can lead to more explainable and accountable AI systems.

\section{Ethnics Statement}
There are no ethics-related issues in this paper. The data and resources utilized in this work are open-source and widely used in many existing studies.

% Entries for the entire Anthology, followed by custom entries
\bibliography{custom}

\newpage
\appendix
\section{Sample Prompts}~\label{sample_prompts}
This section includes some examples of MP under zero-shot and few-shot settings.

\noindent \textbf{Example 1: WiC}
\begin{itemize}
\item \textit{Zero-shot MP Template}: In two sentences, sentence 1: “[sentence\_1]” and sentence 2: “[sentence\_2]”, both contain the target word “[word]”. Determine if the target word is used with the same meaning in both sentences. As you perform this task, follow these steps:
\begin{enumerate}
\item[1.] Understand the context and meaning of the target word in both sentences.
\item[2.] Make a preliminary judgment on whether the target word has the same meaning in
both sentences.
\item[3.] Critically assess your preliminary analysis. If you are unsure about your initial
judgment, reassess it.
\item[4.] Confirm your final answer and explain the reasoning behind your decision.
\item[5.] Evaluate your confidence (0-100\%) in your analysis and provide an explanation for this confidence level.
\end{enumerate}
Provide the answer in your final response as “The target word has the same meaning in
both sentences: \{\} (True / False)”.
\item \textit{Few-shot MP Demonstration}: \\ \textbf{Q:} In two sentences, sentence 1: “The washing machine won't go unless it's
plugged in.” and sentence 2: “The day went well until I got your call.”, both
contain the target word “go”. Determine if the target word is used with the same
meaning in both sentences. \\

\textbf{A:} 
\begin{enumerate}
\item[1.] In sentence 1, “go” refers to the operation of the washing machine. In
sentence 2, “went” (a form of “go”) refers to the progression of the day.
\item[2.] Initially, “go” in the first sentence seems to be about mechanical operation, while in the
second, it's about the passage of time or events. 
\item[3.] Reflecting further, the contexts clearly differentiate the meanings: one deals with machinery and its function, the other deals with the experience of a day. 
\item[4.] The target word “go” does not have the
same meaning in both sentences. In the first, it pertains to operation, and in the
second, to the progression of events. 
\item[5.] I am 83\% confident in this analysis. The
term “go” is used in different contexts in the two sentences, leading to different
meanings. 
\end{enumerate}
The target word has the same meaning in both sentences: False.
\end{itemize}

\noindent \textbf{Example 2: QNLI}

\begin{itemize}
\item \textit{Zero-shot MP Template}: For the question: “[question]” and statement: “[sentence]”,
determine if the statement provides the answer to the question. If the statement contains
the answer to the question, the status is entailment. If it does not, the status is
not\_entailment. As you perform this task, follow these steps:
\begin{enumerate}
\item[1.] Clarify your understanding of the question and the context sentence.
\item[2.] Make a preliminary identification of whether the context sentence contains the
answer to the question.
\item[3.] Critically assess your preliminary analysis. If you feel unsure about your initial
entailment classification, try to reassess it.
\item[4.] Confirm your final answer and explain the reasoning behind your choice.
\item[5.] Evaluate your confidence (0-100\%) in your analysis and provide an explanation
for this confidence level.
\end{enumerate}
Provide the answer in your final response as “The status is {} (entailment /
not\_entailment)”.

\item \textit{Few-shot MP Demonstration}: \\ \textbf{Q:} For the question: “Who was the first President to address the NAACP?” and
statement: “In front of 10,000 people at the Lincoln Memorial, the president left
no doubt where he stood on civil rights.”, determine if the statement provides the
answer to the question. If the statement contains the answer to the question, the
status is entailment. If it does not, the status is not\_entailment. \\
\textbf{A:} 
\begin{enumerate}
\item[1.] The question is asking about the identity of the first President to address the
NAACP. The statement describes a scenario where a president addressed a large
audience at the Lincoln Memorial and took a clear stance on civil rights.
\item[2.] The initial assessment is that the status is not\_entailment.
\item[3.] Upon reassessment, the statement does not specifically mention the NAACP or provide information about which president is being referred to. While the context does pertain to civil rights,
it does not necessarily correlate with the specific act of addressing the NAACP.
\item[4.] The statement does not provide a clear answer to the question. The context
discusses a president's stance on civil rights in front of a large audience at the
Lincoln Memorial, but it does not specify which president this was or confirm that
this president was the first to address the NAACP.
\item[5.] I am 85\% confident in this
analysis. The context lacks specifics about which president is being discussed and
does not directly address the NAACP aspect of the question. 
\end{enumerate}
The status is not\_entailment.
\end{itemize}

\noindent \textbf{Example 3: DDI}
\begin{itemize}
\item \textit{Zero-shot MP Template}: Given the context sentence “[sentence]”, identify the relationship between the pharmacological substances “[item\_1]” and “[item\_2]” within the sentence. Classify the relationship under one of these categories: Advice, Effect, Mechanism, or Int. As you perform this task, follow these steps:
\begin{enumerate}
\item[1.] Understand the context and the meanings of the two substances, along with their potential interactions.
\item[2.] Make a preliminary identification of the relationship between two substances.
\item[3.] Critically assess your preliminary analysis. If you feel unsure about your initial relationship, try to reassess it.
\item[4.] Confirm your final answer and explain the reasoning behind your decision.
\item[5.] Evaluate your confidence (0-100\%) in your analysis and provide an explanation for this confidence level.
\end{enumerate}
Provide the answer in your final response as “The relationship between two substances is {} (Advice / Effect / Mechanism / Int)”.

\item \textit{Few-shot MP Demonstration}: \\ \textbf{Q:} Given the context sentence “Impaired renal function has been described in bone marrow transplant patients who were conditioned with high-dose intravenous melphalan and who subsequently received cyclosporin to prevent graft-versus-host disease.”, identify the relationship between the pharmacological substances “melphalan” and “cyclosporin” within the sentence. Classify the relationship under one of these categories: Advice, Effect, Mechanism, or Int. \\
\textbf{A:} 
\begin{enumerate}
\item[1.] The sentence involves bone marrow transplant patients who were treated with high-dose intravenous melphalan and subsequently received cyclosporin. Melphalan is a chemotherapy drug, and cyclosporin is an immunosuppressive agent. The potential interaction seems related to the effects these drugs have when used together.
\item[2.] The relationship between melphalan and cyclosporin appears to be one of a combined effect where their concurrent use impacts renal function. This suggests the category “Effect”.
\item[3.] Upon reassessment, the relationship still aligns best with the “Effect” category, as the sentence directly connects impaired renal function to the use of both drugs in a sequential treatment regime.
\item[4.] The relationship is classified as “Effect”. The sentence describes a specific medical consequence (impaired renal function) resulting from the combined use of melphalan and cyclosporin in a specific patient group (bone marrow transplant patients), indicating a direct effect relationship between the two substances.
\item[5.] I am 90\% confident in this
analysis. The context provides a clear connection between the two drugs and a specific health outcome, making “Effect” the most suitable category.
\end{enumerate}
The relationship between two substances is Effect.

\end{itemize}

\end{document}